\documentclass[aps,prl,twocolumn,superscriptaddress,amsmath,amssymb, longbibliography, nobibnotes]{revtex4-2}

\usepackage{booktabs}
\usepackage{graphicx}
\usepackage{bibunits} 

\makeatletter
\newcommand{\bibunitcontrol}{%
  \@startbibunitorrelax
  \if@filesw
    \immediate\write\@bibunitaux{\string\citation{apsrev42Control}}%
  \fi
}
\makeatother
\usepackage{hyperref} 
\usepackage{xcolor}

\makeatletter
\newcommand{\apptocfile}{atoc}

\newcommand{\setupSM}{%
  \clearpage
  \onecolumngrid 
  
  \setcounter{page}{1}
  \renewcommand{\thepage}{S\arabic{page}}
  
  \setcounter{figure}{0}
  \renewcommand{\thefigure}{S\arabic{figure}}
  \setcounter{table}{0}
  \renewcommand{\thetable}{S\arabic{table}}
  
  \setcounter{secnumdepth}{2} 
  \setcounter{section}{0}
  \setcounter{subsection}{0}
  \renewcommand{\thesection}{\Alph{section}}
  \renewcommand{\thesubsection}{\Alph{section}.\arabic{subsection}}
  
  \renewcommand{\p@subsection}{}
  \renewcommand{\p@subsubsection}{}
  
  \numberwithin{equation}{section}
  \renewcommand{\theequation}{\Alph{section}\arabic{equation}}
  
  \let\apptoc@orig@addtocontents\addtocontents
  \long\def\addtocontents##1##2{%
    \def\apptoc@ext{##1}%
    \def\apptoc@toc{toc}%
    \ifx\apptoc@ext\apptoc@toc
      \apptoc@orig@addtocontents{\apptocfile}{##2}%
    \else
      \apptoc@orig@addtocontents{##1}{##2}%
    \fi
  }%
}
\newcommand{\appendixtableofcontents}{%
  \vspace{2em}
  \begin{center}
    \textbf{\MakeUppercase{Supplementary Materials}}
  \end{center}
  \vspace{1em}
  
  \@starttoc{\apptocfile}
  
  \vspace{1em}
  \hrule
  \vspace{2em}
}
\makeatother

\hypersetup{
    colorlinks=true,
    linkcolor=blue,
    filecolor=magenta,      
    urlcolor=blue,
    citecolor=blue,
}

\begin{document}

\title{Fault-tolerant foundation models}

\author{Trevor McCourt}
\email{tmccourt@mit.edu}
\affiliation{Department of Electrical Engineering and Computer Science, MIT}

\author{Ila R. Fiete}
\affiliation{Department of Brain and Cognitive Sciences, McGovern Institute, MIT}

\author{Isaac L. Chuang}
\affiliation{Department of Electrical Engineering and Computer Science, MIT}
\affiliation{Department of Physics, MIT}
\date{\today}
\begin{abstract}

Emerging computer hardware often trades reliability for energy efficiency; here we show that large-language models (LLMs) can be trained to tolerate this unreliability, and that rather than degrading, their error resilience actually \emph{increases} as they grow. Modified neural scaling laws inferred from 40,000 GPU-hours of training runs on simulated faulty digital hardware quantify this trend and suggest that models learn to compute within ``good'' error-correcting codes, whose relative overhead remains finite no matter how large the model gets. This finding leads us to conjecture that appropriately trained LLMs may be formally fault-tolerant; if true, running AI inference on low energy, faulty hardware may be a path to substantial energy savings over the status quo.

\end{abstract}

\maketitle

\begin{bibunit}[apsrev4-2]

\bibunitcontrol

Today’s AI systems demand perfectly reliable hardware, and that reliability costs energy. The chips that run commercial AI models \cite{sze2020efficient, jouppi2017datacenter} suppress variability between individual logic gates by operating at high voltage \cite{pelgrom1989matching}; the energy required to toggle a logic gate's state scales with voltage squared. This energy–reliability tradeoff is generic to any physical computing device and central to the thermodynamics of computation \cite{parrondo2015thermodynamics}.

\begin{figure}
\centering
\includegraphics[width=0.75\linewidth]{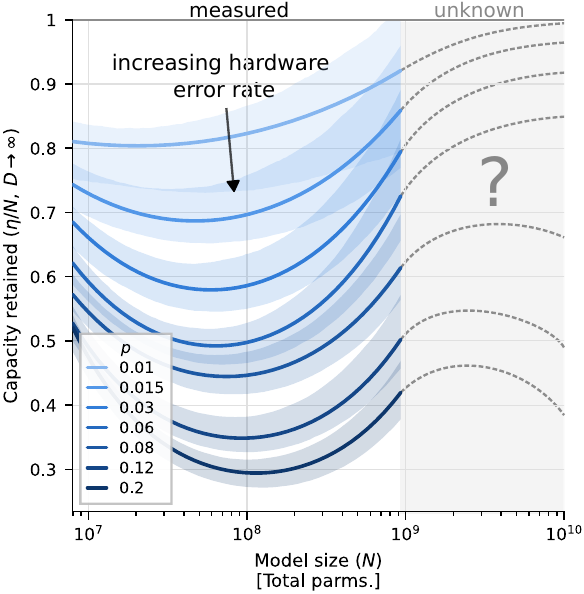}
\caption{
\textbf{The emergence of fault tolerance in AI models as they grow.}
Each curve shows how much {\em useful capacity} a language model retains when trained and run on hardware that introduces errors into arithmetic operations with probability $p$ (darker, more faults). Such a fault-hardened model with $N$ weights performs as well as a conventional model with $\eta$ weights run on clean hardware, so the useful capacity $\eta/N$ is the fraction of the model doing useful work rather than correcting errors (Eq.~\ref{eq:eff_cap}; unlimited training data). Capacity first falls with size, then recovers beyond a critical size $N^*$, driven by a term in the scaling law that fault-free models lack (Eq.~\ref{eq:scaling_law}). This is reminiscent of ``good’’ error-correcting codes, whose overhead stays a fixed fraction at any scale, and we conjecture by analogy that $\eta/N$ approaches a nonzero constant whenever $p$ lies below a threshold (Eq.~\ref{eq:ft_threshold}). Solid curves and shaded bands show measured scaling-law fits and their bootstrap interquartile ranges; dashed curves show possible extrapolations, which depend on how much faults raise the best performance a model can ever attain (Eqs.~\ref{eq:as_collapse} and~\ref{eq:as_sat}).
}
\label{fig:headline}
\end{figure}

Fortunately, biology shows that this reliance on pristine components (and its energy penalty) is not fundamental: the brain, built from heterogeneous, noisy neurons \cite{mainen1995reliability, faisal2008noise}, leverages fault-tolerant circuit architectures to perform complex reasoning and control with orders of magnitude less energy than any existing artificial neural network. Grid cells, for example, compute within an error-correcting code: position is represented redundantly across modules of different spatial periods, so noise in any one module is inconsistent with the rest and can be corrected \cite{sreenivasan2011grid}. Downstream, the grid–hippocampal loop feeds back corrections that cancel slowly accumulating error \cite{burak2009accurate, agmon2020theory, chandra2025episodic}.

AI could be far more energy efficient if run on faulty hardware: analog \cite{yao2020fully, rao2023thousands, ambrogio2023analog}, low-voltage digital \cite{mead_neural, enz2006charge, wang2006sub, mccourt_reliable_arithmetic}, spiking neuromorphic \cite{kudithipudi2025neuromorphic, ornes2025neuromorphic}, photonic \cite{ahmed2025universal, xu2024large}, and probabilistic \cite{borders2019integer, jelinvcivc2026efficient, freitas2026taming} processors all trade absolute reliability for drastic gains in energy efficiency.

However, energy-efficient emerging hardware cannot yet be leveraged for AI at scale because commercial models are fragile: the advanced capabilities \cite{eloundou2024gpts} of large language models (LLMs) often do not survive even quantization (rounding weights to fewer bits) \cite{yu2025superweightlargelanguage, dettmers2023case, dettmers2022gpt3}. Quantization, which perturbs every weight in a deterministic and bounded way, is far gentler than what real low-energy, fault-prone hardware presents (random errors of potentially arbitrary magnitude).

With energy a main bottleneck on AI deployment, developing models that can tolerate hardware errors is among the most important unsolved problems in AI. Computing reliably with unreliable components goes back to von Neumann \cite{vonneumann56}, but was largely set aside once digital hardware became essentially error-free. Energy-constrained AI has revived it: by taking cues from the brain’s grid code, Zlokapa et al. \cite{zlokapa2024fault} recently proved that neural networks can be constructed to compute any function reliably in the presence of errors (although not necessarily efficiently).

The most pressing question concerning fault-tolerant neural network utility is whether models can be efficiently adapted to be insensitive to hardware errors at any size. AI models tend to get larger and more complicated over time, so it is critical to develop techniques that imbue deep learning systems with fault tolerance that improves (or at least does not worsen) as models grow. Encouragingly, some fault tolerance has been built into particular billion-weight models \cite{rasch2023hardware, buchel2026analog}, but only at a single scale: we do not know whether that behavior is part of a persistent trend or sits at the edge of a cliff beyond which errors consume models.

Coding theory suggests that the scaling behavior of fault-tolerance in AI systems ought to be sharply defined. Nature protects information in two ways. The crude way is repetition: keep many copies of a value, as a cell does with duplicated genes, or as simple codes in the brain’s sensory periphery do by spreading information over large populations. The powerful way is a distributed, ``good" code \cite{gallager1962low, sipser1996expander, arikan2009channel} such as what is used in the brain’s grid cells, in which the redundancy needed stays a fixed fraction of the whole no matter how large it grows \cite{sreenivasan2011grid}. If models can learn good codes, a finite fraction of the model performs useful computation at any scale; if they can only manage repetition, that fraction shrinks to zero as models grow. It is not obvious a priori that good codes can be built into transformer models by design or learned by training.

Here we take a critical first step toward running large language models on energy-efficient, unreliable hardware: we show that, unlike conventionally trained models, LLMs trained with the same faults they will later encounter at inference become increasingly error-resilient as they scale (Fig.~\ref{fig:headline}). These results are consistent with models adopting ``good’’ codes rather than simple repetition-based schemes; fault tolerance may therefore persist beyond the scales accessible to our experiments.

The trends in Fig.~\ref{fig:headline} arise because \emph{fault-hardened models} (trained with the same faults they will meet in use) obey slightly different neural scaling laws than conventionally trained, \emph{fault-blind ones}. Neural scaling laws \cite{hestness2017deep, kaplan2020scaling, hoffmann2022training} are the allometry of AI: just as metabolic rate follows a power law in body mass, a model’s prediction error falls as a power law in its number of weights and in its amount of training data. As we detail below, the scaling law of fault-hardened models gains an extra term that lets them improve faster than fault-blind ones at large scale.

We measure computational capacity by $\eta$, the size of a fault-blind model (trained and evaluated without faults) that is required to match the performance of a given fault-hardened model (trained and evaluated with faults),
\begin{equation}\label{eq:eff_cap}
L \left(0, \eta, D \right) = L \left(p, N, D\right)
\end{equation}
where $L \left(p, N, D\right)$ is the loss (prediction error on held-out text) of a model with $N$ parameters whose components fault at rate $p$, trained on $D$ tokens (word fragments).

The relative capacity $\eta/N$ is the neural network analog of a code’s rate: the fraction of the fault-hardened model’s parameters doing useful work rather than absorbing errors. The curves in Fig.~\ref{fig:headline} fit performance data from the thousands of fault-hardened models we trained, indicating $\eta/N$ in the converged limit ($D \to \infty$). The shape of these curves is derived from our newly identified scaling laws,
\begin{equation}\label{eq:eta_scaling}
\begin{gathered}
\ln \frac{\eta}{N} = -c_0 \left( p \right) - c_1 \left( p \right) n + c_2 \left( p \right) n^2 - \mathcal{A}(p, N) \\
n = \ln \frac{N}{N_0}, \: \: N^* = N_0 \exp \left( \frac{c_1 \left( p\right)}{2 c_2 \left( p \right)} \right)
\end{gathered}
\end{equation}
where $N_0$ is an arbitrary size scale and the $c_k$ are simple functions of our scaling law coefficients (Eqs.~\ref{eq:scaling_law} and ~\ref{eq:eta_scaling_constants}). $\ln \eta/N$ is a parabola in log model size with its minimum at $N^*$; the quadratic term is what allows capacity to recover. $\mathcal{A}$ is an asymptotic term that controls behavior beyond our experimental range and is indistinguishable from $0$ within it.

If models were learning to protect themselves from errors using repetition like schemes, we would expect $\eta/N$ to decrease smoothly to $0$ with $N$; the fact that it instead increases past some critical size $N^*$ strongly suggests that models are learning ``good’’ codes comparable to the brain’s grid code. If this is true, for sufficiently weak errors, $\eta/N$ should asymptotically approach a fixed threshold. We return to this limiting behavior later in the manuscript.

It may be surprising that language models could so readily discover good codes without external guidance; however, this is exactly what the principle of error-correction-enhanced evolvability \cite{mccourt_dyn} predicts: adaptive processes (such as learning or evolution) are drawn towards fault-tolerant solutions because fault tolerance allows for more configurations to be explored without risking catastrophic damage. This work is the largest-scale evidence to date for that principle.

Because fault-hardened models show signs of learning to operate within good codes, we conjecture that they may be formally fault-tolerant, retaining finite capacity in the limit of \emph{infinite} $N$,
\begin{equation}\label{eq:ft_threshold}
\lim_{N, D \to \infty} \frac{\eta }{N} \geq c \:, \: \forall \: p < p_{\text{th}}
\end{equation}
for $p_{\text{th}} > 0$ and $0 < c \leq 1$, where $p_{\text{th}}$ is the fault-tolerance threshold (an analog of Eigen’s error threshold for replicating genomes \cite{eigen1971selforganization}) and $c$ is the asymptotic capacity retained.

Our experiments reached $N \approx 10^9$, which was not enough to resolve $\mathcal{A}$ at commercial scales: establishing whether fault-hardened models satisfy Eq.~\ref{eq:ft_threshold} would require experiments orders of magnitude more expensive, which our results strongly suggest are worth doing.

By contrast, fault-blind models are strikingly fragile: on imperfect hardware their scaling behavior is ill-defined and divergent. This is consistent with recent findings that conventional models depend on a handful of outsized ``super weights’'\cite{yu2025superweightlargelanguage}, single points of failure whose number grows the longer a model is trained\cite{gallego2025hidden}.

\subsection*{The scaling behavior of fault-hardened transformers}

\begin{figure}
\centering
\includegraphics[width=0.7\linewidth]{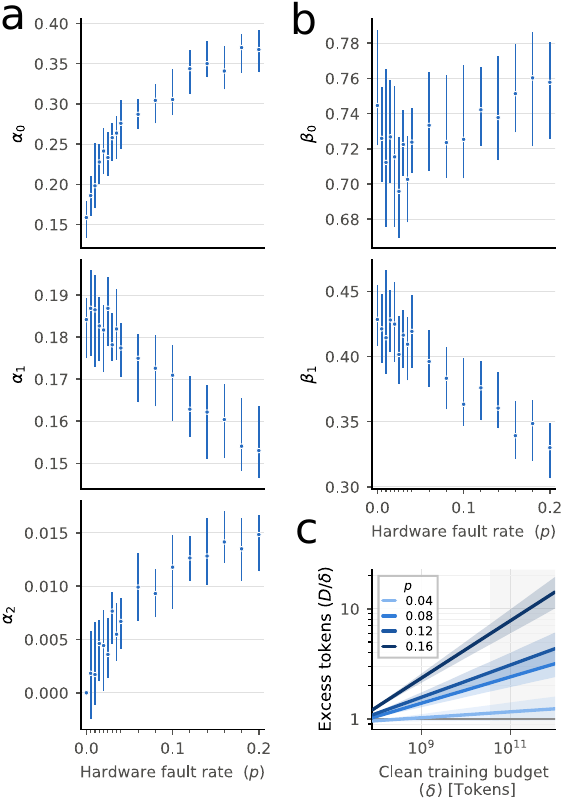}
\caption{\textbf{Training under faults changes how language models scale.}
Fault-hardened transformers obey a scaling law (Eq.~\ref{eq:scaling_law}) with an extra coefficient compared to the standard form ($\alpha_2$); each point fits all models trained at a fault rate $p$ (Methods).
\textbf{a}, Model-size coefficients. $\alpha_1$, the typical model-size coefficient, shrinks as $p$ increases. $\alpha_2$, the curvature term absent from standard laws (pinned at zero for $p=0$), grows with $p$ and drives the recovery of capacity beyond $N^*$ in Fig.~\ref{fig:headline}.
\textbf{b}, Data coefficients. $\beta_1$, the rate at which loss falls with training data, decreases as $p$ increases.
\textbf{c}, Cost of robustness. Tokens $D$ a fault-hardened model needs to match a fault-free model trained on $\delta$ tokens (Eq.~\ref{eq:d_scaling}); gray shading marks extrapolation beyond our largest runs.
Error bars, 95\% confidence intervals; bands in \textbf{c}, interquartile ranges; both from 1,500 bootstrap resamples per $p$ cohort.}
\label{fig:scaling_laws}
\end{figure}

Whether LLMs can tolerate faulty hardware at scale is, like an allometric law, a property of a whole population of models rather than of any single specimen: we established new scaling laws for fault-hardened models by training thousands of them with different $N$, $D$, and $p$ on a slice of FineWeb, a corpus of web text \cite{penedo2024fineweb}.

Our experiments emulated faulty digital hardware that randomly drops blocks of weights out of matrix products. For every forward pass during training or inference, random blocks of $k$ matrix elements within the attention and feed-forward operations of our Llama2-style models were zeroed with probability $p$ \cite{pmlr-v28-wan13}. We fixed $k=4$ throughout and varied $p$ to simulate more or less faulty hardware. Because the faults changed with every query, as synaptic failures change with every spike, fault-hardened models had to learn representations robust to a wide range of failure modes, which is distinct from and harder than calibrating a model to the fixed idiosyncrasies of a specific device \cite{yao2020fully, wright2022deep}, the analog of adapting to a small local lesion. We describe the processor architecture behind this error model below.

Our critical finding was that, under this error model, fault-hardened transformers obey a modified form of the standard scaling law \cite{hoffmann2022training}, with a new curvature coefficient $\alpha_2$ that lets them close the performance gap to their fault-free counterparts as $N$ grows,

\begin{equation}\label{eq:scaling_law}
\begin{gathered}
L(p, N, D) = E \left( p \right) + L_N + L_D \\
L_N \left(p, N \right) = e^{\alpha_0 \left( p \right) - \alpha_1 \left( p \right) n - \alpha_2 \left( p \right) n^2}\\
L_D \left(p, D \right) = e^{\beta_0 \left( p \right) - \beta_1 \left( p \right) d} \\
n = \ln \frac{N}{N_0}, \: d = \ln \frac{D}{D_0}
\end{gathered}
\end{equation}
Here $E$ is the irreducible loss floor, $\alpha_k$ the model-size coefficients, and $\beta_k$ the data coefficients. $N_0$ and $D_0$ are arbitrary scales that render the log arguments dimensionless, both fixed at $4 \times 10^7$.

The restoration of model capacity with size displayed in Fig.~\ref{fig:headline} is driven by the increase of $\alpha_2$ with $p$ (Fig.~\ref{fig:scaling_laws}a), through the relations connecting the $\alpha_k$ and $c_k$ coefficients (Eq.~\ref{eq:eta_scaling_constants}, Methods).

The $p$-dependence of $\alpha_k$ in Fig.~\ref{fig:scaling_laws}a hints at two regimes of fault-hardened scaling, reminiscent of crossing a fault-tolerance threshold. At small $p$, decreases in $\alpha_1$ are mirrored by increases in $\alpha_2$, whereas at large $p$, $\alpha_2$ plateaus while $\alpha_1$ continues to shrink.
Fault-hardened models appear to learn to correct increasingly severe errors up to a point, before being overwhelmed, much as an induced stress response strengthens with the insult until the response itself is overwhelmed.

The asymptotic behavior of $\eta/N$ is controlled by $\mathcal{A}$, which contributed negligibly at the scales accessible to our experiments: shifts in $E$ affect effective capacity only once they are of similar order to $L_N$, and $L_N/E$ fell no lower than $\sim 0.5$ even for our largest models. We return to the limiting behavior in the final section.

\begin{figure*}
\centering
\includegraphics[width=\linewidth]{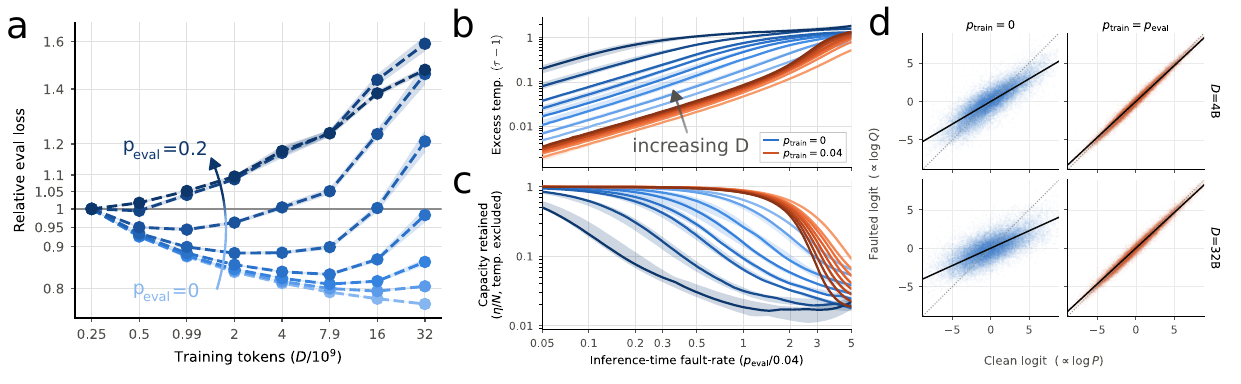}
\caption{
\textbf{Inference-time faults are fatal to models trained without them.}
{\bf a,} The divergence of fault-blind model scaling under inference-time errors. Conventionally, we expect language model performance to increase monotonically with $D$. However, when $p_{\text{eval}} > 0$, fault-blind models do not always exhibit this crucial behavior: performance scaling with $D$ weakens as $p_{\text{eval}}$ increases, and past a threshold, performance can actually decrease with $D$. Shaded regions indicate the standard error of the reported values, estimated from $8$ independently trained models for each value of $D$. \textbf{b,} Error-induced heating of model logits. Part of the degradation caused to models by inference-time errors is heating of the output logits: fault-blind model predictions rise in temperature rapidly with $p_{\text{eval}}$, more so at large $D$, while fault-hardened models resist the heating. \textbf{c,} Error-induced scrambling of model logits. Even after removing the temperature contribution, fault-blind models lose capacity rapidly as $p_{\text{eval}}$ rises above $0$, faster at larger $D$; fault-hardened models maintain almost all capacity for $p_{\text{eval}} < p_{\text{train}}$, independent of $D$. Bands in (b) and (c) indicate the IQR of the reported values, taken over the different trained models and contexts. \textbf{d,} Faulted predictions plotted against fault-free predictions, overlaid for many validation-set inputs. A temperature shift skews the logit blob: the $p_{\text{train}} = 0$ heating is obvious, along with substantial randomization, while the $p_{\text{train}} = p_{\text{eval}}$ logits are barely displaced.}
\label{fig:temp}
\end{figure*}

The data coefficients in Fig.~\ref{fig:scaling_laws}b reflect an additional fact about fault-hardened scaling: models get more expensive to train as $p$ increases (Fig.~\ref{fig:scaling_laws}c). Robustness in artificial neural networks, as in biology, is not free \cite{wagner2005robustness}. Let $\delta$ be the effective training budget of a faulty model, satisfying,
\begin{equation}
L_D(p, D) = L_D(0, \delta)
\end{equation}
Under Eq.~\ref{eq:scaling_law}, $D/\delta$ takes the form,
\begin{equation}\label{eq:d_scaling}
\ln \frac{D}{\delta} = \frac{\beta_0 \left( p \right) - \beta_0 \left( 0 \right)}{\beta_1 \left( 0 \right)} + \left( 1 - \frac{\beta_1 \left( p\right)}{\beta_1 \left( 0\right)}\right) d
\end{equation}
Since $\beta_1 \left( p\right)$ decreases with $p$, the second term's coefficient is positive and increases with $p$.

Since our experiments were capped at $N \approx 10^9$, we did not achieve losses close to the irreducible floor and could not uniquely identify $E$ against the other fit parameters; we therefore fixed $E \approx 1.5$ in all fits (Methods), a choice with no qualitative impact on our results.

Indeed, while Eq.~\ref{eq:scaling_law} is useful for smoothing and interpreting our experiments, it is not strictly necessary for our main conclusions: the trends in Fig.~\ref{fig:headline} are fit-independent and can be seen in the (appropriately processed) raw data; see Supplementary Information Sec.~\ref{sec:raw}.

\subsection*{Fault-blind models collapse under inference-time errors}

In contrast to fault-hardened models, fault-blind models, trained in a sterile, fault-free environment, learn fragile internal representations that fall apart under hardware errors. Fault-hardened models are trained at the fault rate they see at inference, $p = p_{\text{train}} = p_{\text{eval}}$; fault-blind ones see $p_{\text{train}} = 0$ but can still be evaluated with $p_{\text{eval}} > 0$.

The effect of inference-time faults on fault-blind models is not subtle: at sufficiently large $p_{\text{eval}}$, fault-blind models lack convergent scaling laws altogether. As shown in Fig.~\ref{fig:temp}a, when $p_{\text{eval}}>0$, fault-blind transformer performance can actually decrease with $D$, and the value of $D$ at which performance stops increasing seems to fall with $p_{\text{eval}}$.

This catastrophic degradation is consistent with prior observations of super weights/activations in conventional transformers (features of outsized magnitude critical to a model’s functionality) \cite{yu2025superweightlargelanguage}. From the perspective of error-resilience, super-features are single points of failure, the antithesis of the degeneracy that makes biological networks robust \cite{edelman2001degeneracy}. A fault hitting a super-weight, or landing in the path of a super-activation, can degrade a model’s outputs to complete nonsense. Curiously, super-feature counts have been shown to grow with training duration \cite{gallego2025hidden}. This may be directly linked to our degraded scaling laws.

Part of the reason fault-blind models degrade as $p_{\text{eval}}$ increases is that the effective \emph{temperature} ($\tau$) of their output distribution rises. The term is borrowed from statistical mechanics and measures randomness, not heat: at high $\tau$, predictions spread diffusely over possible next tokens $x_t$ rather than concentrating on the most likely one,
\begin{equation}\label{eq:softmax}
\mathbb{P} \left(x_t = k | x_{t-1} = k_{t-1}, \dots, x_0 = k_{0}\right) = \frac{e^{a_k / \tau}}{\sum_j e^{a_j / \tau}}
\end{equation}
where the \emph{logits} $a_k$ depend explicitly on the context $\left( k_{t-1}, \dots, k_0 \right)$. Eq.~\ref{eq:softmax} is the divisive normalization familiar from sensory neuroscience \cite{carandini2012normalization}, with $1/\tau$ playing the role of neural gain.

The effective temperature of fault-blind models rises rapidly as $p_{\text{eval}}$ is raised from $0$ (Fig.~\ref{fig:temp}b), while fault-hardened ones hold a constant temperature until $p_{\text{eval}}$ exceeds $p_{\text{train}}$: fault-hardened models are \emph{climate-controlled}, preserving the precision of their predictions under noise. The heating worsens for both model classes as $D$ increases.

If the only effect of faults were to raise the temperature, the solution would be simple gain control: rescale the logits as a function of $p$ to restore $\tau$ to 1.

\begin{figure}
\centering
\includegraphics[width=\linewidth]{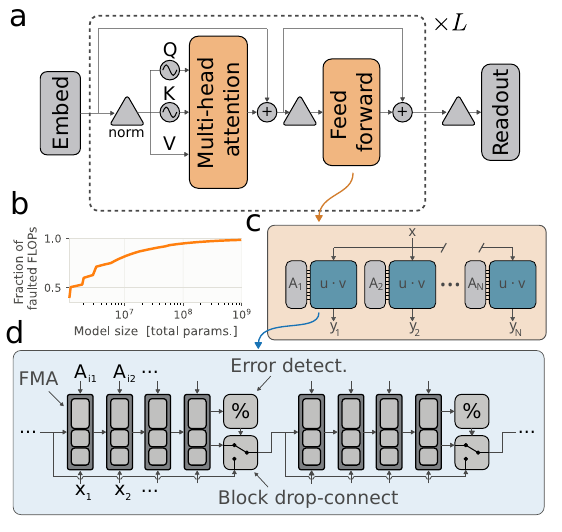}
\caption{\textbf{Error-detecting hardware turns arbitrary arithmetic faults into dropped connections. a,} The operations that faults targeted in our experiments.
We injected errors into the matrix multiplications that drive
the attention and feed-forward operations (orange) in a Llama-2-style transformer.
\textbf{b,} The fraction of fault-targeted operations vs model size. Matrix multiplications (matmuls) account for almost all of the
floating-point operations (FLOPs) required to run transformers; for large models,
more than $ 99\%$ of the FLOP budget.
\textbf{c,} Matrix products reduce to dot products. In hardware, each matmul breaks down into a series of dot products
$u \cdot v$.
\textbf{d,} The error model. A chain of fused multiply–add (FMA) cells accumulates each dot product. Operands carry a redundant residue (their remainder modulo small bases), and after every $k = 4$ cells a cheap modulus check (\%) tests the running sum for consistency. A failed check discards those $k$ products and carries the previous partial sum forward, equivalent to zeroing a block of $k$ elements in one operand; blocks fail independently with probability $p$. Detection thus converts hardware errors of arbitrary magnitude into well-behaved block drop-connect, leaving correction to the model.}
\label{fig:hw_model}
\end{figure}

However, the temperature shift is accompanied by a significant irreversible scrambling of the logits. To be precise, the KL-divergence $D\left(\cdot || \cdot \right)$ between the clean logit distribution $P$ and the faulted distribution $Q$ can be broken into two components,
\begin{equation}\label{eq:kl_split}
D \left( Q | P \right) = D \left( Q | P^* \right) + D \left( P^* | P \right)
\end{equation}
where $P^*$ is the nearest distribution to $Q$ that is a strict temperature shift of $P$ (Methods).

The first term in Eq.~\ref{eq:kl_split} is the part of the gap between $P$ and $Q$ that no rescaling can close; Fig.~\ref{fig:temp}c plots the capacity lost to this term as $p_{\text{eval}}$ increases. Notably, fault-hardened models lose almost no capacity until $p_{\text{eval}}$ exceeds $p_{\text{train}}$, while fault-blind ones deteriorate rapidly from $p_{\text{eval}} = 0$. Fig.~\ref{fig:temp}d plots clean logits against their faulted counterparts, clarifying the distinct contributions of temperature rescaling vs.\ scrambling.

\subsection*{Low-energy neural network accelerators}

Our fault model emulated a Low Energy Neural Network Accelerator (LENNA) \cite{mccourt_reliable_arithmetic}: a chip that runs its arithmetic at voltages too low for perfect reliability, but detects its own errors and converts each one into a dropped connection (the artificial counterpart of a synapse that fails to transmit).

Our experiments were consistent with a hardware system in which LENNA cores execute the primary transformer matrix products (in the attention and feed-forward layers), while all the supporting operations (nonlinearities, norms, positional encoding, and embedding/readout layers) are run on some other fault-free substrate; see Fig.~\ref{fig:hw_model} (a). As shown in Fig.~\ref{fig:hw_model} (b), matrix products overwhelmingly dominate the FLOP budget of scaled transformer models.

LENNAs build error detection into matrix products by encoding operands within a redundant residue number system (RRNS) \cite{rao1974error, barsi2006error}, where extra ``check moduli" are carried through a matrix product so the correctness of results can be verified \cite{parhami1999computer}. This is the same principle by which grid cells encode position: each module reports location modulo its spatial period, and redundancy across modules exposes errors \cite{sreenivasan2011grid}. In hardware, matrix–vector products are broken into dot products (Fig.~\ref{fig:hw_model}c) computed by chains of fused multiply–add (FMA) circuits. In a LENNA, each FMA also processes the residues, and after every $k$ FMAs, a check compares the residues with the primary result; if they disagree, those $k$ operations are discarded, and the previous intermediate result is carried forward (Fig.~\ref{fig:hw_model}d).

Importantly, LENNAs only detect errors; correcting them is left to the model it runs. This division of labor is efficient because, for RRNS arithmetic, detection is far cheaper than correction. An $N$-bit (array-style) FMA uses $\propto N^2$ full adders; in a 12-bit example with $k=4$, carrying an extra 3-bit residue adds only a few percent, and the consistency check needs only residue-sized adders, negligible beside the FMAs. Full correction in hardware requires an algorithm based on the Chinese remainder theorem that could easily exceed the cost of the FMAs themselves. The arrangement echoes the brain’s, where grid modules expose errors cheaply, and correction is delegated to the downstream grid–hippocampal loop \cite{burak2009accurate, sreenivasan2011grid}.

\subsection*{Towards formally fault-tolerant AI}

The ultimate objective of fault-hardened neural network research should be to establish that Eq.~\ref{eq:ft_threshold} holds for realistic hardware fault models: that fault-hardened models running on actual faulty hardware retain capacity at arbitrary scale, as the brain evidently does. Practically, it would suffice to show that at some large enough scale (e.g., $10^{11}$ parameters), $\eta/N$ stays large enough for a fault-hardened model on low-energy faulty hardware to beat a fault-blind one on an existing accelerator by a margin that justifies the change of hardware.

If our scaling law holds asymptotically, we expect one of two behaviors, with $p$ controlling a smooth transition between them. If the floor-gap $\Delta$ is much larger than the remaining size-loss $L_N$ at scale, the capacity retained by the fault-hardened model collapses to $0$,
\begin{equation}\label{eq:as_collapse}
\ln \frac{\eta}{N} \propto -n
\end{equation}
Alternatively, if $\eta/N$ approaches $1$ while the gap remains small compared to $L_N$, then $\mathcal{A}(p, N)$ becomes a negative correction that slows the growth of $\eta$, leading to saturation near $N$,
\begin{equation}\label{eq:as_sat}
\mathcal{A} \left( p, N\right) \approx \frac{\Delta \left( p\right)}{\alpha_1 \left( 0\right) L_N \left( p, N \right)}
\end{equation}

The value of $p$ at the boundary between collapse and saturation may be operationally defined as the fault-tolerance threshold at a given experimental scale, much as the error threshold of a replicating genome separates heritable information from mutational meltdown \cite{eigen1971selforganization}.

\section*{Discussion}

We have identified additional terms in the scaling laws for fault-hardened transformers that lead to large models performing surprisingly well (relative to what would be expected via simple error-correcting schemes like repetition).

To determine whether this behavior persists at the $\sim 10^{11}$ weight scale, as our work hints is possible, models must be trained in that regime. We estimate that a reasonable extension of our experiment would cost around $10^{26}$ FLOPs, or $\sim 10^8$ GPU-hours (similar to the total training budget of Llama-3 405B \cite{grattafiori2024llama}), which is intractable within academia.

Smaller experiments could de-risk the large one. Simple natural or synthetic datasets, where the asymptotic regime is easily reached, can serve as model organisms: fault-hardened transformers trained on them could be studied as a function of the dataset’s irreducible floor $E$. Fault tolerance that exists at small scale and persists as $E$ increases would be strong evidence that commercial-scale experiments are worth doing.

Further evidence that fault-hardened models learn good codes may be found in the structure of their weights and activations. Understanding how large neural networks work is notoriously hard \cite{jonas2017could, lipton2018mythos}. Still, there is precedent: networks trained to navigate spontaneously develop grid-cell-like codes \cite{cueva2018emergence,banino2018vector}, and the stark contrast between how fault-blind and fault-hardened models respond to faults (Fig.~\ref{fig:temp}) suggests an equally detectable difference in their internals. Our trained models, multiple terabytes of weights and optimizer states, are available to anyone who wants to look (see Data availability).

The brain is the existence proof that fault tolerance, energy efficiency, and intelligence are compatible, and the principle of error-correction-enhanced evolvability \cite{mccourt_dyn} holds that fault tolerance is an expected outcome of adaptation in a noisy environment. Our results suggest that AI trained under the constraints that shaped nervous systems may come to share their robustness. Additionally, hardware built with error-detection codes, as found in the brain, may provide the substrate for bounded-error, low-energy accelerator operation.

Establishing the commercial viability of fault-hardened networks will require substantial further investment, but the reward is commensurate: on future low-energy hardware, they could be orders of magnitude more energy efficient than any inference system existing today.

\section*{Methods}

\subsection*{Dataset}

Every model was trained on a 350-billion-token slice of FineWeb~\cite{penedo2024fineweb}, tokenized with an 8192-token byte-pair-encoding vocabulary and packed into 1024-token sequences. The corpus was shuffled in large blocks, so any contiguous slice is statistically equivalent to any other. Performance was measured on a held-out split of two billion tokens, drawn from the same web crawls but sharing no documents with the training data: near-duplicates were removed by a normalized-content hash, and the tokenizer was fit on training documents alone.

\subsection*{Model ladder}

All models are Llama2-style decoder-only transformers (rotary position embeddings, RMSNorm, and SwiGLU feed-forward blocks), following a standard reference implementation~\cite{karpathy2023llama2c}. The ladder holds every architectural ratio fixed and varies only overall size: the attention head dimension is 32 at every rung (head count is width over 32, no grouped-query attention), the width-to-depth ratio is held near 32, the context length is 1024 tokens, and the embeddings are untied. Sixteen logarithmically spaced (width, depth) pairs range from $7.9 \times 10^{6}$ to $9.3 \times 10^{8}$ total parameters; throughout, $N$ refers to total parameters, embeddings included.

One architecture (width 512, depth 13, $N = 5.0 \times 10^{7}$) was additionally trained at eight token budgets spanning two orders of magnitude, with eight random seeds per condition and six training fault rates. These runs provide the fixed-$N$ measurements in Fig.~\ref{fig:temp}.

\subsection*{Fault injection}

Faults were applied to the operands of the matrix products that dominate a transformer’s FLOP budget: the query, key, value, and output projections of the attention block, the two context products $QK^\mathsf{T}$ and $AV$ ($K$ and $V$ being the faulted operands), and the gate, up, and down projections of the feed-forward block. Each faulted operand was masked along its contraction axis in contiguous blocks of $k$ elements, every block zeroed independently with probability $p$, which is exactly the effect of an error-detecting FMA chain discarding its last $k$ accumulations (Fig.~\ref{fig:hw_model} (d)); blocks fail independently for every row of the product. The embedding lookup, readout, norms, rotary embeddings, nonlinearities, and residual additions were left fault-free. A fresh mask was drawn for every sequence on every forward pass and held fixed across positions within it, so a model trains against the whole fault distribution rather than one realization of it. Throughout, $k$ was fixed at 4 and $p$ took seventeen values between 0 and 0.2.

\subsection*{Training}

We used a typical training procedure and ``conservative" hyperparameter choices lifted from a standard reference implementation \cite{karpathy2023llama2c}. More elegant tuning schemes \cite{yang2021tensor, yang2021tuning, dey2026don, mlodozeniec2026completed} slightly improved the performance of large models as expected, but introduced systematic deviations from typical scaling laws that confounded our analysis; since all our measurements are differential across $p$-values rather than concerned with absolute performance, the added complexity was not worthwhile.

The number of training tokens was set per run as a fixed multiple of the parameter count, giving token budgets from $7.9 \times 10^{7}$ to $3.5 \times 10^{10}$. The two axes were not fully crossed: the ten architectures at or below $1.1 \times 10^{8}$ parameters were each trained at 10, 20, 80, and 320 tokens per parameter, while the six larger ones were trained only at 20 and below, since completing the ladder at the top would have cost an order of magnitude more compute than the entire sweep. Every combination of size, duration, and fault rate was trained at two random seeds.

\subsection*{Scaling-law fits}

Equation~\ref{eq:scaling_law} was fit to each $p$ cohort separately, so no assumption about how the coefficients vary with $p$ is built into the estimates. Fits were performed on $\log L$, which weights runs evenly across the orders of magnitude the sweep spans, with a Huber loss to prevent a few off-trend points from dominating, and from multiple starting points because the objective is not convex. The pivots $N_0$ and $D_0$ are a gauge choice rather than fitted quantities, both fixed at $4 \times 10^{7}$.

Two quantities were held rather than fitted: the curvature $\alpha_2$, pinned to zero for the fault-free reference cohort and fit freely for every other cohort, and the floor $E$, fixed at 1.55 for every cohort, the value preferred by the $p=0$ cohort. The choice of $E$ mainly rescales $\alpha_1$: fixing $E = 1.85$ instead gives $\alpha_1 \approx 0.3$ at $p=0$ with no meaningful change to the other coefficients, consistent with the literature \cite{hoffmann2022training, besiroglu2024chinchilla, muennighoff2023scaling}. Deviations from published values are expected given the simpler dataset and smaller tokenizer used here.

Confidence intervals come from bootstrapping: the fitted runs were resampled with replacement 1500 times per cohort, and the surface was refit on each resample. Because the $(N, D)$ grid is a design rather than a sample, a resample that drops the largest models loses the only constraint on the long-$N$ end of the surface and reports its absence as uncertainty in the exponents; the intervals we quote are conservative for that reason.

The capacity relations of Eq.~\ref{eq:eta_scaling} follow from solving $L(0, \eta, D) = L(p, N, D)$ (Eq.~\ref{eq:eff_cap}) for $\eta$ in the converged limit $D \to \infty$. Expressed through the fitted coefficients of Eq.~\ref{eq:scaling_law},
\begin{equation}\label{eq:eta_scaling_constants}
\begin{gathered}
c_0 = \frac{\alpha_0 \left( p \right) - \alpha_0 \left( 0 \right)}{\alpha_1 \left( 0 \right)}, \: \:
c_1 = 1 - \frac{\alpha_1 \left( p \right)}{\alpha_1 \left( 0 \right)}, \: \:
c_2 = \frac{\alpha_2 \left( p \right)}{\alpha_1 \left( 0 \right)} \\
\mathcal{A}(p, N) = \frac{1}{\alpha_1 \left( 0 \right)}\ln \left( \frac{\Delta \left( p \right)}{L_N \left( p, N\right)} + 1\right) \\ \Delta \left( p \right) = E \left( p\right) - E \left( 0 \right)
\end{gathered}
\end{equation}
The asymptotic term $\mathcal{A}$ depends on the floor gap $\Delta$, which is indistinguishable from zero within our experimental range.

\subsection*{Logit-space analysis}

The measurements in Fig.~\ref{fig:temp} were made on a fixed set of 256 contexts, each the final token position of a held-out sequence. For each context and fault rate, we averaged the model’s next-token distribution over 1000 independent fault draws, giving the fault-averaged predictive $Q$; the fault-free distribution $P$ was recorded for the same contexts. This was repeated at 81 fault rates between 0.002 and 0.2.

A single effective temperature was fit per model and fault rate by minimizing $\sum_c D(Q_c | P^*_c)$ over $\tau$, the sum running over contexts and $P^*$ being $P$ rescaled by $\tau$ as in Eq.~\ref{eq:softmax}. Since the rescaled distributions form a one-parameter exponential family in $1/\tau$, the minimizer is the temperature matching the mean logit under $P^*$ to its mean under $Q$, and the split of Eq.~\ref{eq:kl_split} into explained and residual parts is exact for the sum over contexts.

\section*{Declarations}

\subsection*{Data and Code Availability}

The complete codebase used to produce the results presented in this manuscript is publicly available \cite{fht_code}. In particular, this repository contains a set of Jupyter notebooks that generate this manuscript’s figures, along with many other supplementary plots that help support our conclusions. These notebooks rely on summary data \cite{fht_summary_data} which records the performance of our trained models, but omits their weights. Model weights are also available; this dataset is multiple terabytes and may be accessed by running the command \texttt{aws s3 ls s3://ft-experiment-weights/ --request-payer requester} through the AWS CLI.

\subsection*{Author Contributions}
TM conceptualized, designed, executed, and interpreted the results of the experiments conducted in this work. ILC and IF provided advice throughout the project and helped compose the manuscript.

\subsection*{Competing Interests}
All authors declare no competing interests.
\putbib[references]

\end{bibunit}

\setupSM 

\appendixtableofcontents

\begin{bibunit}[apsrev4-2]

\bibunitcontrol


\section{N-curvature in the raw data}\label{sec:raw}

The specific functional form chosen for the N-dependence of Eq.~\ref{eq:scaling_law} has no bearing on the conclusions presented in Fig.~\ref{fig:headline}. The recovery of performance at large scales is visible in the raw data after subtraction of the $D$-dependence. 

\begin{figure*}[h]
    \centering
    \includegraphics[width=\linewidth]{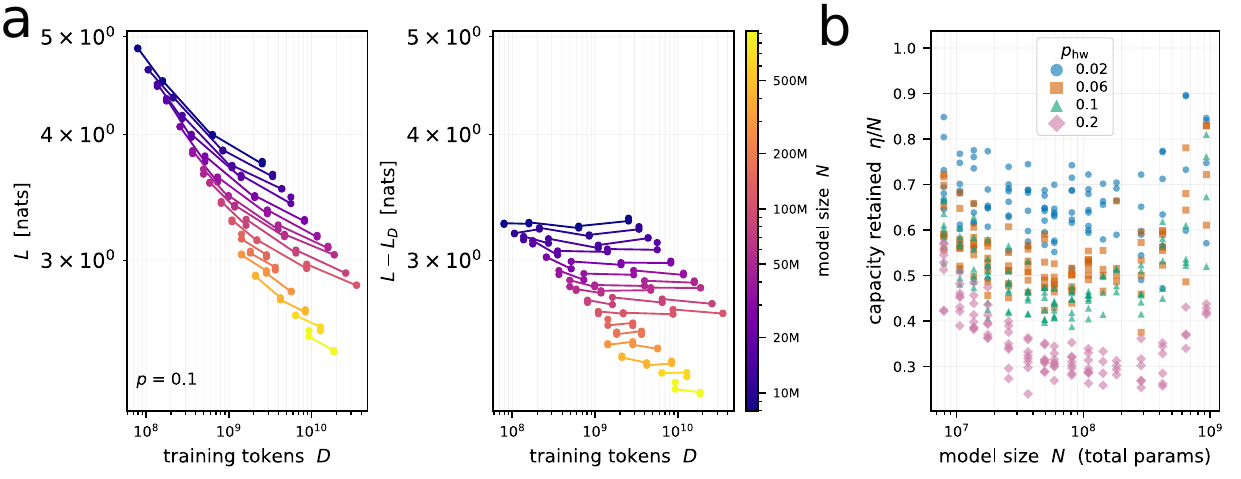}
        \caption{\textbf{Retained-capacity signals are visible in the raw data. (a)} Subtracting the $D$-dependence from the raw data. The left panel shows the raw experimental data; the right panel shows the same data after $L_D$ has been subtracted from each loss value. Essentially all $D$-dependence is removed from the data, as far as is visible within run-to-run variation. We show the subtraction for a single value of $p$; the effective datasets for all values of $p$ are given in a notebook included in the code accompanying this manuscript \cite{fht_code}. \textbf{(b)} Retained-capacity signals in the $L_D$-subtracted dataset. Once every datapoint has been brought to convergence by removing the $D$-dependence, the standard Chinchilla law fitted to the $p=0$ cohort can be used to compute $\eta$ for each point in the data.}
    \label{fig:sm_raw}
\end{figure*}

To reveal the trends shown in Fig.~\ref{fig:headline} in the raw data, we fit the scaling laws to each $p$-cohort as usual to identify the parameters $\beta_0 \left( p \right)$ and $\beta_1 \left( p \right)$, and then subtract $L_D$ from every point in the original dataset to effectively bring each one to convergence. This transformation is shown in Fig.~\ref{fig:sm_raw} (a): the leftmost plot shows the experimentally collected $(L, N, D)$ data for a particular value of $p$, and the rightmost plot shows the $(L - L_D, N)$ data produced by the subtraction. As can be seen from the flatness of the contours in the rightmost plot, the subtraction entirely removes the $D$-dependence from the data (within the bounds of run-to-run variation).

After the subtraction is complete, we can use the standard Chinchilla scaling law identified for the $p=0$ cohort to assign a value of $\eta$ to each point in the effective dataset. Fig.~\ref{fig:sm_raw} (b) shows the results of this projection: exactly the same curvature shown in Fig.~\ref{fig:headline} is shown in this figure.

\section{Residuals of the scaling law fit}

\begin{figure*}[h]
    \centering
    \includegraphics[width=\linewidth]{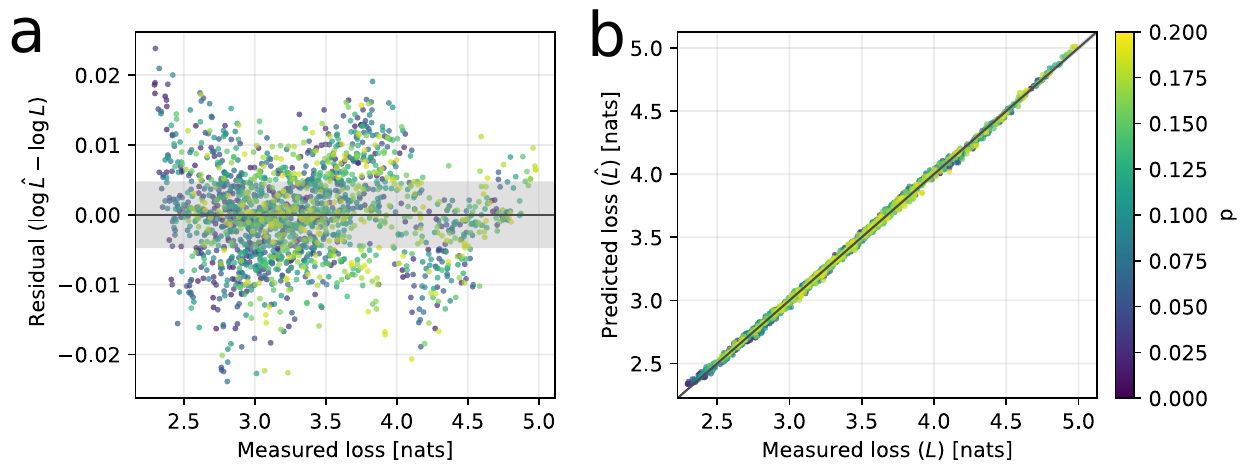}
    \caption{\textbf{ Residuals of the scaling law fit (a)} The log-residuals of our scaling law fit ($\hat{L}$) vs measured loss ($L$), for all measured values of $p$. The scale of the residuals is near what would be expected due to run-run variation alone (indicated by the gray band). \textbf{(b)} The loss predicted by our model vs the measured loss. The black line indicates a perfect fit.}
    \label{fig:sm_resid}
\end{figure*}

The ultimate reason that the trends shown in Fig.~\ref{fig:headline} are not corrupted by the choice of the $N$-dependence in the scaling law is that the scaling law fits the data well. The goodness-of-fit of Eq.~\ref{eq:scaling_law} to our data is demonstrated by the plots shown in Fig.~\ref{fig:sm_resid}. 

\section{The necessity of the curvature term}

In contrast to the precise fit of our new law demonstrated in Fig.~\ref{fig:sm_resid}, fits of the standard Chinchilla form (without the new curvature term) to $p>0$ cohorts fail miserably. 

\begin{figure*}[h]
    \centering
    \includegraphics[width=0.6 \linewidth]{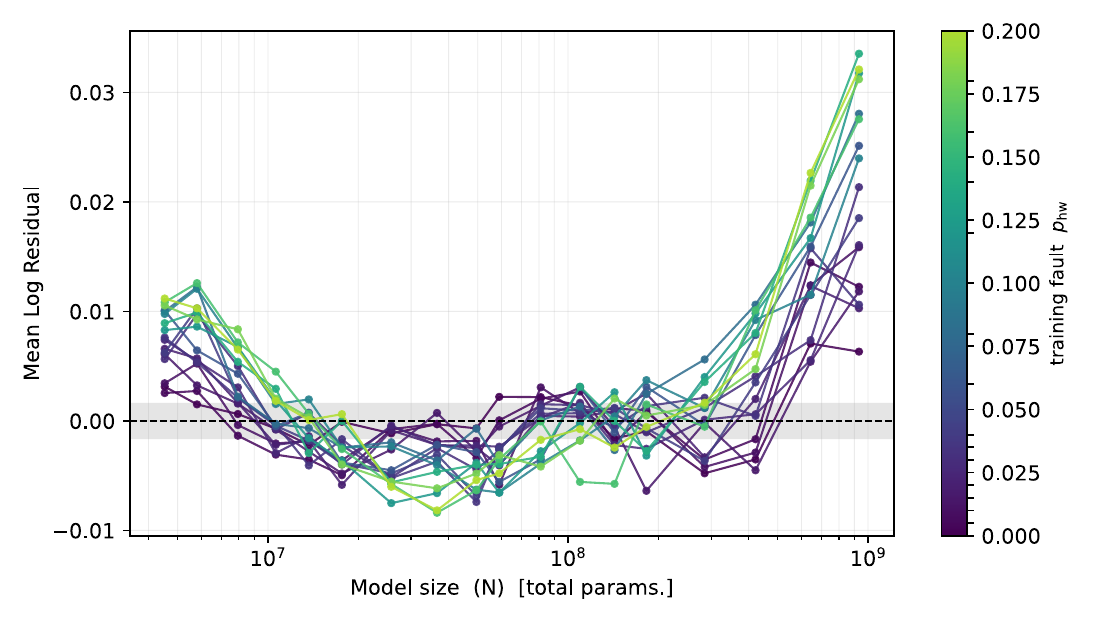}
    \caption{\textbf{ Residuals in the scaling law fit without the curvature term} Residuals in a fit of the standard Chinchilla form to our datasets. The curvature term is clearly needed to represent the trends in the raw data.}
    \label{fig:sm_std}
\end{figure*}

Fig.~\ref{fig:sm_std} shows the log residual of the curvature-free fit vs $N$, averaged over different values of $D$ and independent training runs. For $p>0$, there is a clear curvature in the residuals which is ultimately what inspired our addition of $\alpha_2$ to the law.

\section{Correlations between scaling law parameters}

We quantified the uncertainty in our scaling law fit parameter values via bootstrapping. We drew many synthetic datasets (1500) by resampling the experimental runs with replacement and refit the scaling law to each one. This procedure left us with many samples of possible scaling law parameter sets, which approximate the true distribution of fit parameters given the data. The error bars shown in Fig.~\ref{fig:scaling_laws} were generated by estimating 95\% confidence intervals from our bootstrap samples.

\begin{figure*}[h]
    \centering
    \includegraphics[width=\linewidth]{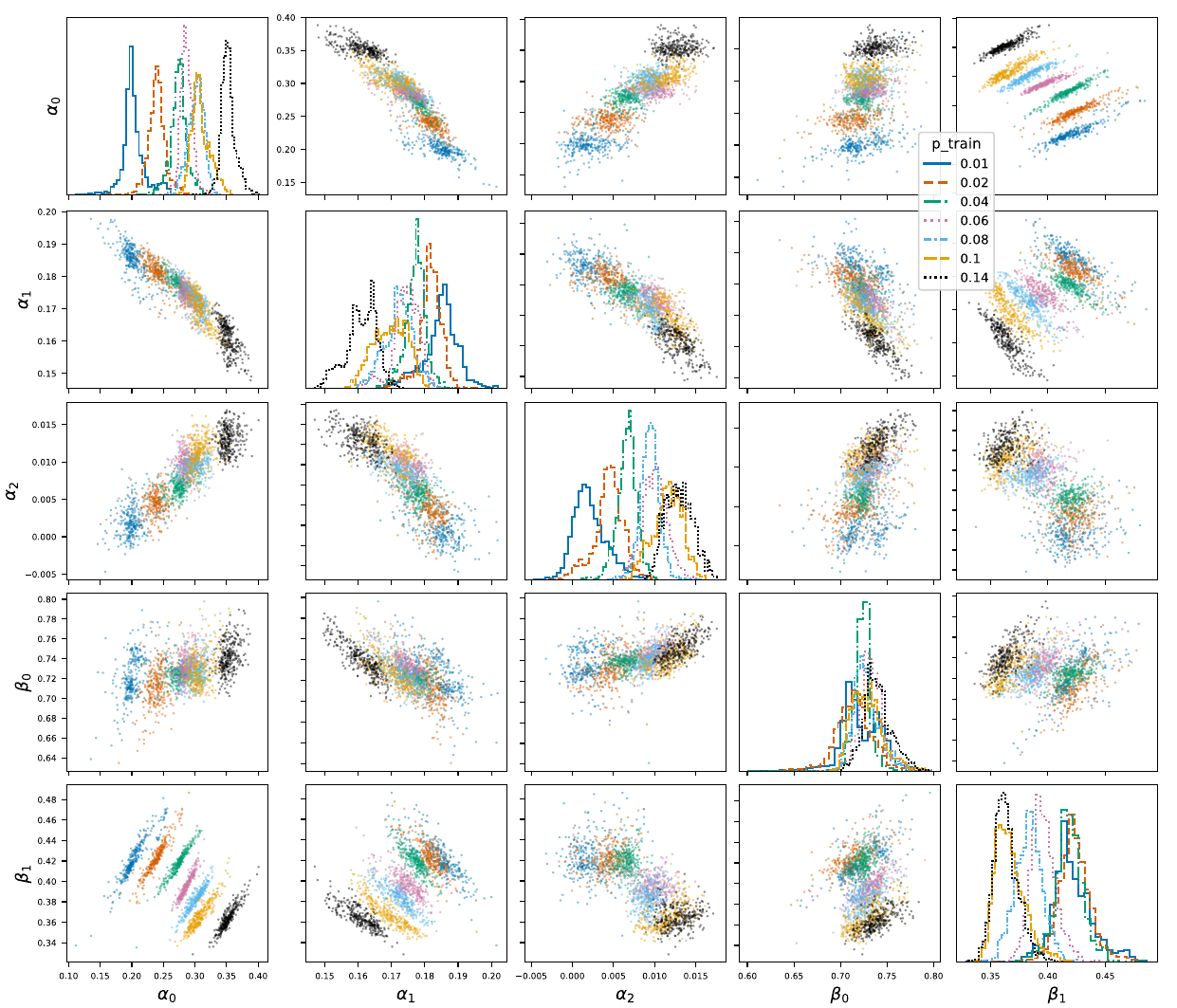}
    \caption{\textbf{Pairwise correlations between scaling law parameters} Pairwise scatter plots that identify correlations between scaling law parameters. A small subset of the complete bootstrap dataset is shown to avoid overcrowding the plots with points.}
    \label{fig:sm_corr}
\end{figure*}

The parameter samples generated by our bootstrapping procedure are useful for revealing correlations between the different fit parameters, as shown in Fig.~\ref{fig:sm_corr}.

While there are weak correlations between the lower order $N$ and $D$ coefficients, $\alpha_2$ is essentially entirely uncorrelated with any other fit parameter. This makes sense, as no other parameter in Eq.~\ref{eq:scaling_law} is capable of producing the curvature shown clearly in Fig.~\ref{fig:sm_std}.


\end{bibunit}

\end{document}